\documentclass[fleqn,usenatbib]{rasti}

\usepackage{newtxtext,newtxmath}

\usepackage[T1]{fontenc}

\DeclareRobustCommand{\VAN}[3]{#2}
\let\VANthebibliography\thebibliography
\def\thebibliography{\DeclareRobustCommand{\VAN}[3]{##3}\VANthebibliography}

\usepackage{graphicx}	
\usepackage{amsmath}	
\usepackage{booktabs}
\usepackage{xspace}
\usepackage{xcolor}

\newcommand{\astronomaly}{\textsc{Astronomaly}\xspace}
\newcommand{\name}{{\tt{AHUNT}\,}}
\newcommand{\namefs}{{\tt AHUNT}}

\newcommand{\OracleEpoch}{round~}

\title[Human in the loop anomaly detection]
{Learning to Detect Interesting Anomalies}

\author[A. Vafaei Sadr  et al.]{
Alireza Vafaei Sadr ,$^{1,2}$\thanks{E-mail:}
Bruce A. Bassett,$^{3,4,5}$
Emmanuel Sekyi$^{3,4}$
\\
$^{1}$Département de Physique Théorique and Center for Astroparticle Physics, University of Geneva\\
$^{2}$Institute of Pathology, University Hospital RWTH Aachen, Pauwelsstraße 30, 52074, Aachen, Germany\\
$^{3}$African Institute for Mathematical Sciences, 6 Melrose Road, Muizenberg, 7945, South Africa\\
$^{4}$Department of Maths and Applied Maths, University of Cape Town, Cape Town, South Africa\\
$^{5}$South African Astronomical Observatory, Observatory, Cape Town, 7925, South Africa
}

\date{Accepted XXX. Received YYY; in original form ZZZ}

\pubyear{2015}

\begin{document}
\label{firstpage}
\pagerange{\pageref{firstpage}--\pageref{lastpage}}
\maketitle

\begin{abstract}
Anomaly detection algorithms are typically applied to static, unchanging, data features hand crafted by the user. But how does a user systematically craft good features for anomalies that have never been seen? Here we couple deep learning with active learning -- in which an Oracle iteratively labels small amounts of data selected algorithmically over a series of rounds -- to automatically and dynamically improve the data features for efficient outlier detection. This approach, \namefs, shows excellent performance on MNIST, CIFAR10, and Galaxy-DESI data, significantly outperforming both standard anomaly detection and active learning algorithms with static feature spaces. Beyond improved performance, \name also allows the number of anomaly classes to grow organically in response to the Oracle's evaluations. Extensive ablation studies explore the impact of Oracle question selection strategy and loss function on performance. We illustrate how the dynamic anomaly class taxonomy represents another step towards fully personalized rankings of different anomaly classes that reflect a user's interests, allowing the algorithm to learn to ignore statistically significant but uninteresting outliers (e.g. noise). This should prove useful in the era of massive astronomical datasets serving diverse sets of users who can only review a tiny subset of the incoming data. 
\end{abstract}

\begin{keywords}
Anomaly detection -- Human in the loop -- Deep learning
\end{keywords}



\section{Introduction}

Anomaly, novelty, and outlier detection are the interrelated tasks of separating the exotic from the normal; the strange from the standard. Almost all commonly used anomaly detection algorithms - isolation forest, local outlier factor, OC-SVM and density estimation methods, to name just a few - are data agnostic: they work on any data features given to them by a user, and in turn, do not change or alter those features. This generality is a powerful benefit, but it has significant downsides too. The performance of any anomaly detection algorithm on a given set of data may vary from excellent to terrible. 

Anomalies, or at least anomalies of interest to a user, may be effectively invisible in the features supplied by the user of the algorithm and the algorithm can do nothing about this. As a trivial example, consider an astronomer whose data includes  unknown transients with time-varying brightness. If the features chosen by the astronomer do not include any time dependence, no anomaly algorithm will be able to separate the transients from the static background objects. This is highly reminiscent of machine learning before deep learning, in which feature engineering was critical, and largely separate, from the algorithms themselves. However, whereas standard feature engineering is tricky, at least the user has insights from the raw data as a guide. In the case of anomaly detection there are often no examples of anomalies beforehand, and from the scientific view, this is exactly the most interesting situation since it opens the door to completely new discoveries. The entire undertaking becomes significantly harder when the datasets become exponentially larger, since human eyes look at less and less of the data. This is precisely the situation we face in astronomy today where future transient surveys will yield massive datasets with highly complex selection effects, see e.g. \cite{plasticc, lsst_optimal}. 

To mitigate these problems a number of approaches have been used. First, one can explore different anomaly algorithms and optimise the myriad of associated hyperparameters and automatically generate different classes of data features (see e.g. \cite{sadr2021flexible}). However, there are no assurances that this will successfully detect anomalies \footnote{The celebrated No Free Lunch theorems mean that no algorithm can detect anomalies better than random if there are no limits on the type of anomaly and no other prior information \citep{NFL}.}.

A more sophisticated approach is to incorporate dynamic human feedback about whether objects are anomalous or not through {\em active learning}. In active anomaly learning an Oracle is  used to provide labels for a small number of selected examples that allows the algorithm to iteratively improve (see e.g. \cite{pelleg2004active}, \cite{nixon2021salad}, \cite{pimentel2018generalized}, \cite{activelearning-01}). Although the initial selection of potential anomalies may be mostly uninteresting, through active learning the hope is to progressively learn to focus on the interesting anomalies.  

Recent examples of active anomaly learning for astronomy is the {\tt Astronomaly} algorithm \citep{lochner2021astronomaly, astronomaly_hitch} and \cite{Walmsley_2022}. In {\tt Astronomaly} the feedback from the human Oracle  is used to learn a function over the data features that reflects the interests of the user. This function is then blended with the raw anomaly score to learn which parts of the feature space are more interesting to the user, allowing a personalised ranking of potential anomalies that can learn to exclude highly anomalous but boring examples (e.g. pure noise), as illustrated in \cite{astronomaly_apply}. 
This approach learns which parts of the feature space are interesting to the scientist, but does not alter the underlying features in any way. Hence, if the underlying features chosen by the user are poor, even this active learning, human-in-the-loop, feedback will be of little help in finding the anomalies of interest. To return to our example of transient detection, no amount of human feedback will be able to imbue the static features with information about the time dependence of the sources. 

This standard approach to anomaly detection is what we will call ``{\em static}" anomaly detection with active learning: the features are supplied by the user and are unchanging and frozen in time. In contrast, in this paper we extend the Active Learning framework of {\tt Astronomaly} to allow the human feedback to be used to dynamically evolve the features to make them  progressively more and more relevant for the anomalies of interest to the user. Of course, this Active Learning approach cannot perform miracles. If examples of the anomalies are never presented to the human Oracle for labelling the algorithm will never get the chance to learn better features. 

At this point the reader may rightly ask why we don't simply apply Deep Learning to the problem; and indeed there have been many papers doing just this. However, although deep learning solves the feature engineering problem for standard machine learning problems, it does not resolve the core problem stated above. Deep learning famously requires large amounts of data \citep{aggarwal2018neural} and learns features adapted for the majority class; exactly the opposite regime than the one of interest. These majority features may well provide good features for detecting anomalies but they may equally well not. To return to the transient toy model example, the deep learning algorithm may learn visual shape filters ideal for distinguishing point sources from spiral galaxies, but would not learn temporal features needed to detect transients. 

In this paper we combine active and deep learning in \name (Anomaly Hunt) and explore its ability to overcome all of these various limitations of current approaches. The outline of the rest of this article is as follows. Section 2 includes an introduction to the \name algorithm. We give a brief overview of the datasets and evaluation metrics are given in Section 3. In Section 4, the results are presented and discussed in detail, and we conclude with a summary and a short discussion in Section 5.

\section{Overview of \name} 

Our stated goal is to create an algorithm that dynamically changes the  data features initially input by the user to make them increasingly better suited for detecting anomalies. Further, we want the algorithm to adapt preferentially to the anomalies of most interest to the user, not just to anything that is anomalous. This is important since extremely noisy, and hence anomalous, but uninteresting examples, such as image artefacts, are common in real-world systems. 
Active learning, where the user is the Oracle, kills both birds with one stone. 

To learn adaptive features useful for anomaly detection we need three components: 
\begin{itemize}
    \item An anomaly detection algorithm that ranks objects based on how anomalous they are in the current feature set.
    \item An Oracle question selection algorithm that presents the Oracle with examples to be labelled as interesting or not.  
    \item A method to augment and use the Oracle feedback to evolve and adapt the data features to make them more useful for detecting anomalies. 
\end{itemize}
 
In this paper we choose Deep Learning to adapt our features. Deep learning, and Convolutional Neural Networks (CNN) in particular, have the key ability to learn relevant features for the classification or regression task at hand. As discussed before, this does not help much in the standard anomaly detection setting since the learned features are dominated by the normal classes since anomalies are, by definition, rare. Hence, whether the anomalies stand out with respect to these learned ``normal" latent space features is purely a matter of chance. We wish to guide this learning process to ensure that the learned features are adapted for anomaly detection. We use our model to rank order anomalies. We then display the first $n_{\rm anomalies}$ to the Oracle for evaluation. This feedback is used to retrain the model to improve the representations learned in the feature space.

In principle we could use any optimisation algorithm to perform the final step above of adapting the features. For example, we could use a genetic programming algorithm that learns suitable nonlinear functions of the input data to optimise anomaly detection performance. Or, if the input data is high-dimensional, one could use a genetic algorithm to learn which combinations of features are most useful for detecting anomalies. While these approaches are certainly interesting, and may even be superior in some contexts, deep learning is an attractive starting point and we will leave the exploration of other approaches to future work. This choice informs our nomenclature: we will often, and interchangeably, refer to the feature space that we evolve and which we perform anomaly detection on as the latent space, reflecting the fact that it is often very different from the raw input data features.   While the above description is fully general, in this paper we will focus on visual anomaly detection tasks for the sake of testing the algorithm. However, the approach should work in any setting.  

\begin{figure*}[h]

 \flushleft
\hspace*{-0.35in}
\includegraphics[width=1.1\textwidth, 
]{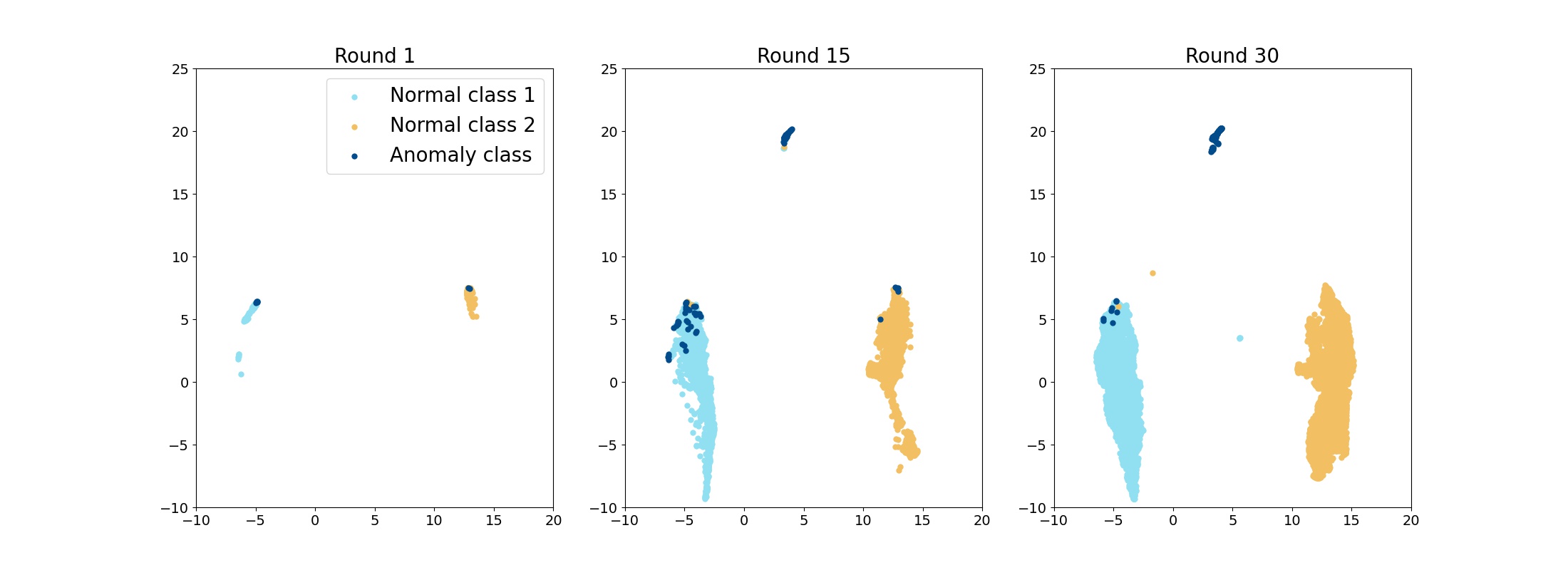}
\caption{Illustration of the \name dynamic latent (feature) space evolution for MNIST over 30 rounds of Active Learning using UMAP dimensionality reductions of the latent space down to two dimensions. In round 1 (left) the anomalies (dark blue) are degenerate with the normal classes (beige and light blue). By round 15 a large separate cluster of anomalies has formed and by round 30 the anomalies are completely separated from the beige cluster. Note that this comes at the cost of dispersing the clusters of normal data, but that is acceptable since our goal is not to classify the normal data.}
\label{evolving_features}
\centering
\end{figure*}


We now discuss the \name pipeline in detail. We assume that at the start we have a labeled dataset with objects classified over $n$ known, "normal" classes\footnote{A natural extension is to consider \name applied to a regression problem in which we predict a relevance score rather than a class. Below we explore multiple subclasses of anomalies where the user can specify their relative interest in the various subclasses, which provides a bridge to the regression case.}; e.g. a set of galaxy images classified as spiral, elliptical or irregular. Further assume that we have no information on, and no examples of, anomalies.   

\name works iteratively over a series of learning iterations that we call {\em Rounds}. These rounds may be defined either by the availability of the Oracle (e.g. if the data is static or streaming), or may correspond to new data being periodically added to the database (e.g. a new night's worth of telescope data).  The oracle is asked to answer a set number of questions, $n_q$,  in each round (i.e. to label $n_q$ examples). The examples are chosen via active learning with the goal of helping the algorithm learn the most informative features for detecting interesting anomalies.  In addition to the $n$ known classes we start with a single "reserve" class for new anomalies (i.e. there are $n+1$ classes in total). 

Before the Oracle answers any questions the algorithm begins by learning to classify the initial data over the $n$ known classes. After the first Round, \name makes predictions over the $n+1$ known+Reserve classes for the new data.  

If the user finds a new class of anomaly they can extend the existing $n$ known classes to include the new anomaly class, meaning there are now $n+2$ classes in total, including the reserve class. This process can be repeated at every round, allowing \name to dynamically expand its classification taxonomy to include an arbitrary number of new and anomalous classes over the rounds. 

After one (or more) new classes is added, the Oracle or user is able to weight the anomaly classes differently based on how interesting they are. This allows the algorithm to pick examples to label that focus more on the interesting anomalies. 

One key factor is how the active learning chooses the questions in each round.  To rank potential questions one can use an anomaly score if a classic anomaly detection algorithm is used. However, the outcome of the deep learning model in \name is a pseudo-probability vector over the classes. One can use this pseudo-probability, $p_{c}$, of the target class $c$, to choose examples to be labelled. 

There are a large array of possible active learning algorithms to use for selecting samples for the Oracle to label. Common choices are uncertainty-based sampling, Bayesian optimisation and expected error minimization. We compare three different approaches: (i) choosing the examples that \name is least confident in its predictions, (ii) choosing the most anomalous examples and (iii) choosing randomly (as a control).  There is significant freedom in implementing these strategies, especially when there are multiple anomaly classes of interest.

Here we describe the simple approach to choosing the most uncertain examples we have taken; it is by no means unique. We ask the user to specify their relative interest over all $n+1$ classes at the current round. We use this to split the $n_q$ Oracle questions between the different classes (e.g. the user may specify 100\% weight on the Reserve class, or 50\% on a known anomaly class and 50\% on the Reserve class). For each class of interest, $c$, we then produce a  list of all unlabelled examples ranked according to an uncertainty index, $\rho_c$ (which lies between $0$ and 1):
\begin{equation}
    \rho_c = 1-2(p_{c}-0.5) 
\label{eq:uncertainty_index}
\end{equation}
Here $p_{c}$ is the probability of the class, $c$, in question and we only consider examples where  $p_{c}>0.5$ to focus on candidates likely to be of the class of interest. As discussed before, this could be replaced by more sophisticated active learning strategies, but suffices for our demonstrations.

The Oracle is then asked to label the requisite number of examples from the top of each list (where the sum of all the examples is $n_q$). Once labeled, these data are augmented so that all classes have the same number of examples as the dominant class and the model is retrained using all of the available labelled training data. This leads to the evolution of the latent space learned by the \name over the rounds of active learning, as shown in Fig. (\ref{evolving_features}), that leads to better separation of anomalies from the normal classes. Initially all anomalies lie amongst the normal classes (blue, beige) which are well separated. Over the various rounds the algorithm learns to separate most of the anomalies from the two normal classes. 

Although our goal in developing \name was to perform well on any active learning problem applied to any base machine learning model, we chose to test it in the context of image classification. As a result, we choose to use the convolutional neural network (CNN) deep learning models for all our datasets. For MNIST and CIFAR-10 we choose a 4-layer network while choosing 6-layer model for DESI. Kernels have size three, and each convolution layer is followed by a pooling layer. We choose a 64-dimensional latent space and use the ADAM optimizer with a learning rate equal to $10^{-5}$ and train the network for three epochs after each round. For most of our results we use a softmax layer as the final output layer. However in Appendix (\ref{sec:loss_function}) we explore the impact of the choice of loss function on algorithm performance, showing that the IsoMax loss function may offer some advantages, especially in very heterogeneous situations.


\subsection{Comparison Algorithms and Ablation Tests} \label{sec:comparison}

To properly test the utility of \name we compare it to existing anomaly detection methods and to static active learning approaches (such as those used in \astronomaly \citep{lochner2021astronomaly}, DRAMA \citep{sadr2021flexible} and Zoobot \citep{Walmsley_2022}). In particular we compare \name to:

\begin{itemize}
        \item{\bf Static Active Learning} - this is our most important comparison since it allows us to determine the value added by the dynamic latent space features. In the static case we use the same \name neural network pretrained on all the data available from round 0 with a reserve anomaly class (and with no anomaly examples as before). The Oracle then answers questions selected as in standard \namefs. These answers are used to adapt only the final layer on top of the pretrained latent feature space. This is our closest analogue to \astronomaly \footnote{Note that the implementation of active learning in \astronomaly is somewhat different since it does not use a neural network as its base. Instead it can work with any set of user-supplied features and then uses the Oracle feedback to learn a  regression model over the static feature space supplied by the user. } and Zoobot and allows the algorithm to learn which parts of the feature space are interesting, but does not deform the feature space. It represents \name with all layers frozen other than the final layer. This is the model called "Static" in the plots, e.g. Fig (\ref{fig:all_three_datasets_dynamic_static_new_2}).


    \item {\bf Random Active Learning} - in this model we allow for dynamic evolution of the latent space as in the full version of \name but randomly selected examples are given to the Oracle at each round, without paying any attention to anomalies or prediction uncertainty. This model allows us to examine the relative impact of the algorithm used to select questions for the Oracle to answer.
    

    \item {\bf Iforest-latent-dynamic} - Here we apply the unsupervised isolation forest anomaly algorithm  \citep{liu2008isolation} directly on our dynamic latent space that comes from retraining the classifier after each round of oracle answers. We therefore expect these results to improve after each round of active learning. This is very similar to \name except that it uses an anomaly detection algorithm on the high-dimensional latent space rather than feeding the results into a final neural network layer.

    \item {\bf Iforest-latent-static} - Here we apply isolation forest to the latent space coming from training the deep classifier only on the 0-th \OracleEpoch dataset, i.e. the classifier is trained on only normal data. This latent space has learned about the important features in classifying the normal classes but knows nothing about the anomalies, is static and does not use active learning at all. We expect it to perform poorly. 
    
    \item {\bf Iforest-raw} - Here we use isolation forest applied to the raw input pixel data rather than the learned latent space. Since the dimensionality of the input data is often very large this algorithm typically struggles. It does not use active learning at all.

    
\end{itemize}

These algorithms allow us to dissect the performance of \name and attribute its success to the various novel components in the algorithm. 

\section{Datasets and Metrics}

We use three datasets to evaluate \name on a range of image domains. The first two are classic  AI datasets (MNIST and CIFAR-10) while the third is the DESI (Dark Energy Spectroscopic Instrument) galaxy image dataset. In each dataset we randomly choose two of the classes as our "normal" classes and another one as the anomaly class. 

\begin{table*}

\begin{tabular}{llllllll}
\toprule
Dataset{} &                                   Normal Classes &           Anomaly Class & \#0-th Round & \#Normal / Round & Anomalies / Round & \#Rounds & \#Questions / Round ($n_q)$ \\
\midrule
MNIST    &                                               0, 1 &                 2 &             500, 900 &         170, 200 &                 6 &       30 &                   5 \\
CIFAR-10 &                               Airplane, Automobile &              Bird &             400, 300 &         140, 150 &                 2 &       30 &                  10 \\
DESI     &  Unbarred Tight Spirals \&  &  Merging  &             500, 400 &           66, 50 &                15 &       20 &                  10 \\
{} &                                   Edge-on  w/o Bulge & {} & {} & {} & {} & {} & {} \\ 
\bottomrule

\end{tabular}
\centering
\caption{Information about the experiments we ran on the three datasets (MNIST, CIFAR-10 and DESI). In each case we give the classes treated as "Normal" and "Anomalies" in each case, as well as the number of objects in each of the normal classes at round 0 (i.e. before active learning starts) and in each subsequent round. We also give the number of anomalies present in each round's data (which varies from about 1-13\% of each round's new data) and the number of questions answered in a round by the Oracle ($n_q)$.  Here \# should be read as "Number of". We choose a limit in which there are both few anomalies and few Oracle questions.}
\label{table_data}
\end{table*}

In each dataset, we simulate a set of time-dependent observations by providing five parameters: the numbers of the two normal classes in the data at round zero and then the numbers of normal and anomaly classes in each successive round (taken to be the same each round for simplicity).  This is a simple simulation of a rolling search typical of modern time-domain search survey. 

The second key component in the observing strategy is the Oracle question strategy. This is  controlled by two factors: (1) the number of questions, $n_q$, put to the Oracle each round and (2) how \name chooses which $n_q$ data points from the data from that round to put forward to the Oracle for labelling, discussed earlier. 
We choose $n_q$ to be comparable, but typically smaller than, the number of anomalies each round. We also chose $n_q$ to be small (either 5 or 10) per round to provide a very challenging environment to test \name in. In practise, for astronomical applications such as LSST, it is reasonable for humans to label upwards of 1000 objects per night, and potentially many more if a citizen science platform like GalaxyZoo \citep{fortson2012galaxy} is leveraged.  See Table \ref{table_data} for all parameters related to the datasets. 

\begin{figure*}

\includegraphics[width=14cm]{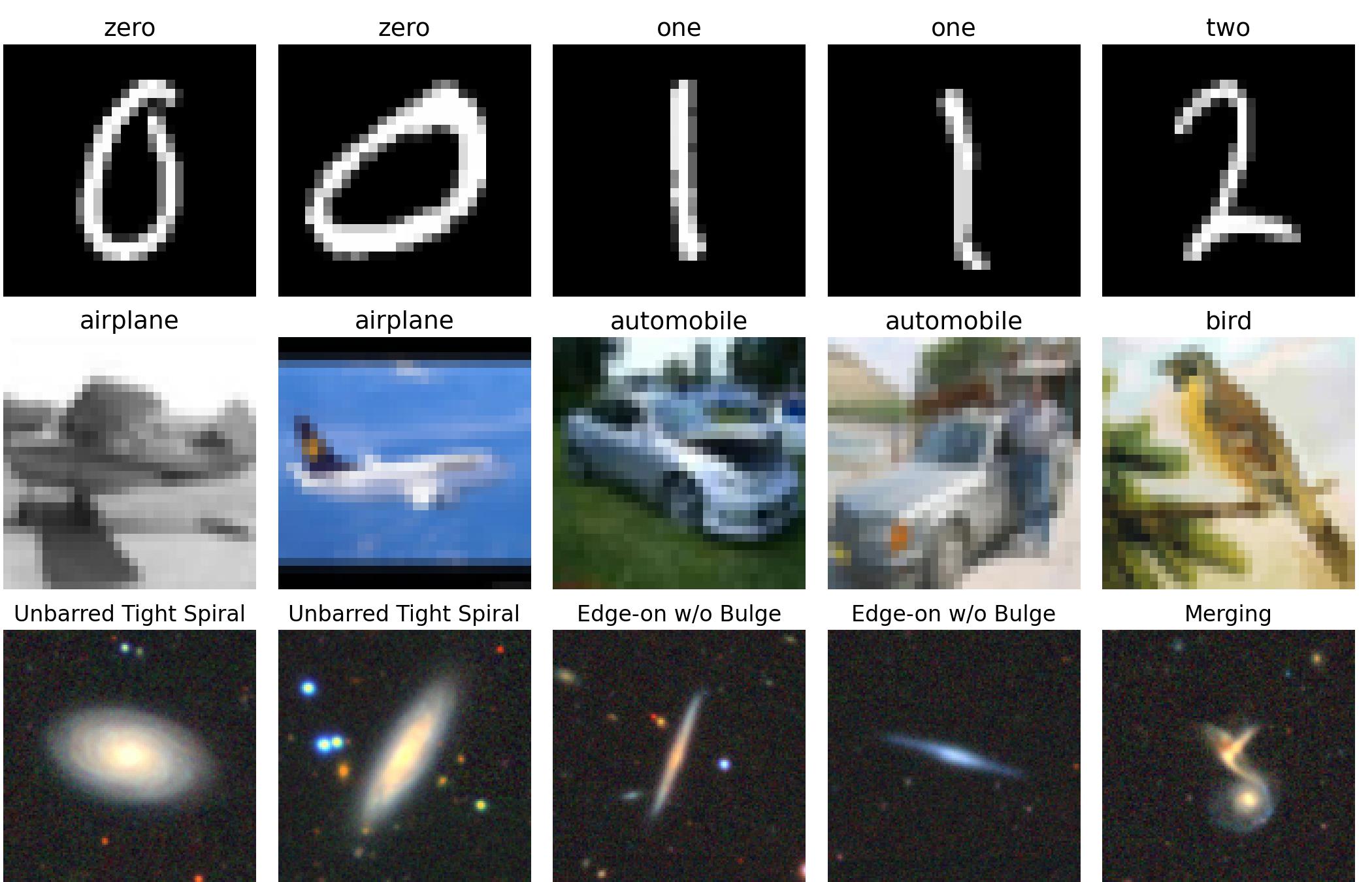}
\caption{Random samples from the three datasets we explore: MNIST (top row), CIFAR-10 (middle row) and DESI (bottom row). The first two columns show examples from the two normal classes we chose while the last column shows an example from the anomaly class. For more information see Table \ref{table_data}.}
\centering
\label{fig:dataset_examples}
\end{figure*}

Here we briefly describe each of the datasets used for our experiments. Examples of all three datasets showing the normal classes (first four columns) and anomalies (fifth column) are shown in Fig. (\ref{fig:dataset_examples}).

\subsubsection*{MNIST}

Our evaluation begins with the well-known MNIST dataset that contains 70,000 handwritten digits resized into $28\times28$ pixels$^2$ images, normalized and centered \citep{deng2012mnist}.  Our active learning experiments start with 500 instances of `0' and 900 cases of `1' as the Normal classes (round 0). Then 170 `0' instances, 200 `1' instances and 6 instances of `2's (the anomaly class) are added in each of the 30 rounds; see Fig. (\ref{fig:dataset_examples}). 5 questions are answered by the Oracle at each round ($n_q = 5$). To augment the data we employed random rotations of up to 10 degrees and up to 4\% width/height shifts and zooms.

\subsubsection*{CIFAR-10}

The second dataset is the well-known CIFAR-10 dataset \citep{krizhevsky2009learning} that consists of 60,000 colour images of size $32\times32$ pixels drawn from 10 classes. This dataset is similar to MNIST but more complex. See \ref{fig:dataset_examples} for examples. We utilized small-size images to explore various configurations of the experiments. 

This experiment begins with 400 instances from the `airplane' class and 300 instances of `automobiles' as the two normal classes. Then 140 `airplane' instances and 150 `automobile' instances, together  with two `bird' examples  (the outliers) are added for each of the subsequent 30 rounds. 10 questions were given to the Oracle each round ($n_q = 10$). As with the MNIST data, we employed rotations and shifts \& zooms of up to 10 degrees and 4\% respectively to perform the augmentations.  

\subsubsection*{DESI}

The final dataset we explored was the astronomical DESI Legacy Imaging Surveys (DESI),  \citep{dey2019overview} where the resolution and image quality are much higher than in the other datasets. The DESI dataset combines three astronomical galaxy datasets and includes 18000 images coming from 10 broad galaxy classes assigned using volunteer voting.  All images are resized to $128 \times 128$ pixels. See \ref{fig:dataset_examples} for examples. 

For the DESI data we take the normal classes to be \textbf{Unbarred Tight Spiral Galaxies} and \textbf{Edge-on Galaxies without Bulge} beginning with 500 and 400 instances respectively. For the 20 rounds of active learning we then provide an additional 66 \textbf{Unbarred Tight Spiral Galaxies} and 50 \textbf{Edge-on Galaxies without Bulge} instances,  together with 15 \textbf{Merging Galaxies} (the outlier). Again 10 questions were asked to the Oracle at each round ($n_q = 10)$. We also employ a 45 degrees rotation and a 5\% width/height shift and zoom as data augmentation for Galaxy10 DESI.

\subsection{Evaluation Metrics}

In this work we primarily tracked metrics suitable for anomaly detection (since metrics such as accuracy are typically very poor in this context). In particular we considered the 
Matthews Correlation Coefficient (MCC), the Rank-Weighted Score (RWS) \citep{badac} and the AUC. 
We found empirically that the results for all three of these metrics were very highly correlated in our case and as a result we report only the results for the MCC metric for simplicity.  
The MCC lies between -1 and 1 \citep{matthews1975comparison} and is defined by:
\begin{equation}
\mbox{MCC} = \frac{TP \times TN - FP \times FN}{(TP+FP)(TN+FN)(TP+FN)(TN+FP)}
\end{equation}

where $T (F)$ stands for ``True'' (``False'') and $P (N)$ for ``positive'' (``negative'') respectively.  
In addition to the MCC we also track the fraction or number of anomalies correctly identified in each \OracleEpoch. See e.g. the ``True Candidate'' panel in Fig. (\ref{fig:multi_anomalies}). An effective active learning strategy will see this quantity rise quickly over the rounds as the algorithm learns the right latent space features that allow easy detection of anomalies.

\section{Results}

To test \name we consider initially a simplified scenario in which there is only one class of anomalies in the data.  \name starts with the two known normal classes and a reserve/outlier class which initially has 100\% of the focus of the active learning algorithm. Once the new anomaly class is discovered the class taxonomy expands to four: two normal classes (of no real interest), the new anomaly class and the reserve/outlier class. Now 100\% of focus for Oracle question selection switches away from the reserve class to the anomaly class. This setup will be used for all of our ablation tests and comparisons. For our metrics and plots we demand that the algorithm must correctly identify examples as belonging to the new anomaly class, not as belonging to the reserve class. 

In section (\ref{sec:dynamic_taxonomy}) we switch to a more realistic and exciting scenario which demonstrates the flexibility of \name in dealing with a dynamic taxonomy/hierarchy of classes: we allow both the number of anomaly classes and the user's relative interest in the different classes to change with time. This is more realistic and allows the user to personalize the algorithm's attention by distributing questions between the known anomaly classes ("known unknowns") and as-yet-undiscovered anomalies ("unknown unknowns").  


\subsection{Dynamic vs Static Features}

How well does the dynamic feature extraction in \name perform in improving  anomaly detection results in this context? We  compare \name performance against a scenario which we call "static". The static feature space only applies transfer learning at each round: namely, the model is fully trained only on data available in the zeroth round, i.e. before seeing any anomalies. Then all layers of the model are frozen except for the last layer which is fine-tuned through active learning. This case is the closest scenario to what \astronomaly proposed: the network learns which regions of the fixed latent space are interesting to the user.

\begin{figure*}
\includegraphics[width=17cm]{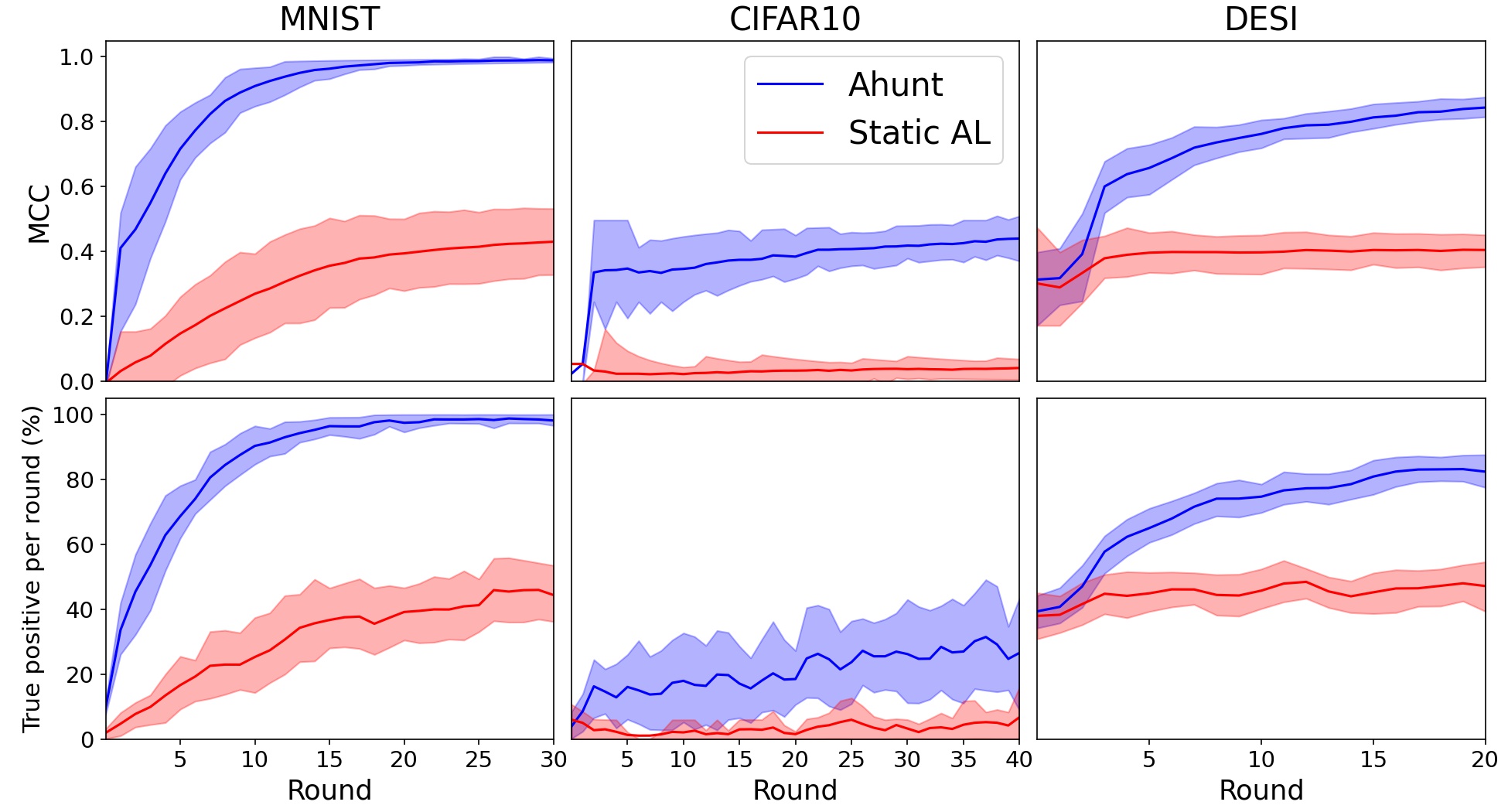}
\centering
\caption{The performance of \name on the three datasets MNIST, CIFAR10 and DESI in comparison to active learning with static latent space features (Static AL). In all cases \name significantly outperforms the static version of the algorithm in which there is no evolution of the latent space from one round of active learning to the next. In the static case only the weights of the final layer connecting  the output probabilities to the (static) latent space are updated. The first row shows the MCC score while the second row of figures shows the fraction of new anomalies correctly identified as anomalies. For MNIST \name rapidly learns to classify all anomalies while for CIFAR10 and DESI the improvement is slower though significantly better than the static case. The CIFAR10 results are dominated by the rarity of the anomalies in this case (only two per round); see Table \ref{table_data}.
Error contours are 68\% regions computed from 50 randomised runs at each round. To compare the performance of \name against all algorithms see Fig. \ref{fig:all_three_datasets_ALL_Algs}.}
\label{fig:all_three_datasets_dynamic_static_new_2}

\end{figure*}

We compare the MCC scores for all our datasets for \name compared with the static (transfer learning) scenario in table \ref{table:AllActive}. We see that having a dynamic feature space adds a significant boost over simple transfer learning, more than doubling the MCC scores at the end of all the rounds. This will not be true always of course: there will be anomaly detection problems for which the advantage is smaller, however notice that transfer learning usually excels when there is little training data, which is exactly the case here; yet evolving the entire network yields superior performance which we attribute to the intelligent choice of examples by the active learning. The performance of the algorithms over each round is shown in Fig. \ref{fig:all_three_datasets_dynamic_static_new_2} which shows how, even after a few rounds, \name quickly adapts to finding good anomaly features which static transfer learning either learns slowly or gets stuck, showing little improvement over time. This possibility was noted in \cite{lochner2021astronomaly} where responsibility for a good choice of features was squarely placed on the user.   These results show that \name can alleviate much - and in some cases all - of this pressure.

\subsection{Active Learning vs Random Question Selection}

\begin{figure*}
\includegraphics[width=17cm]{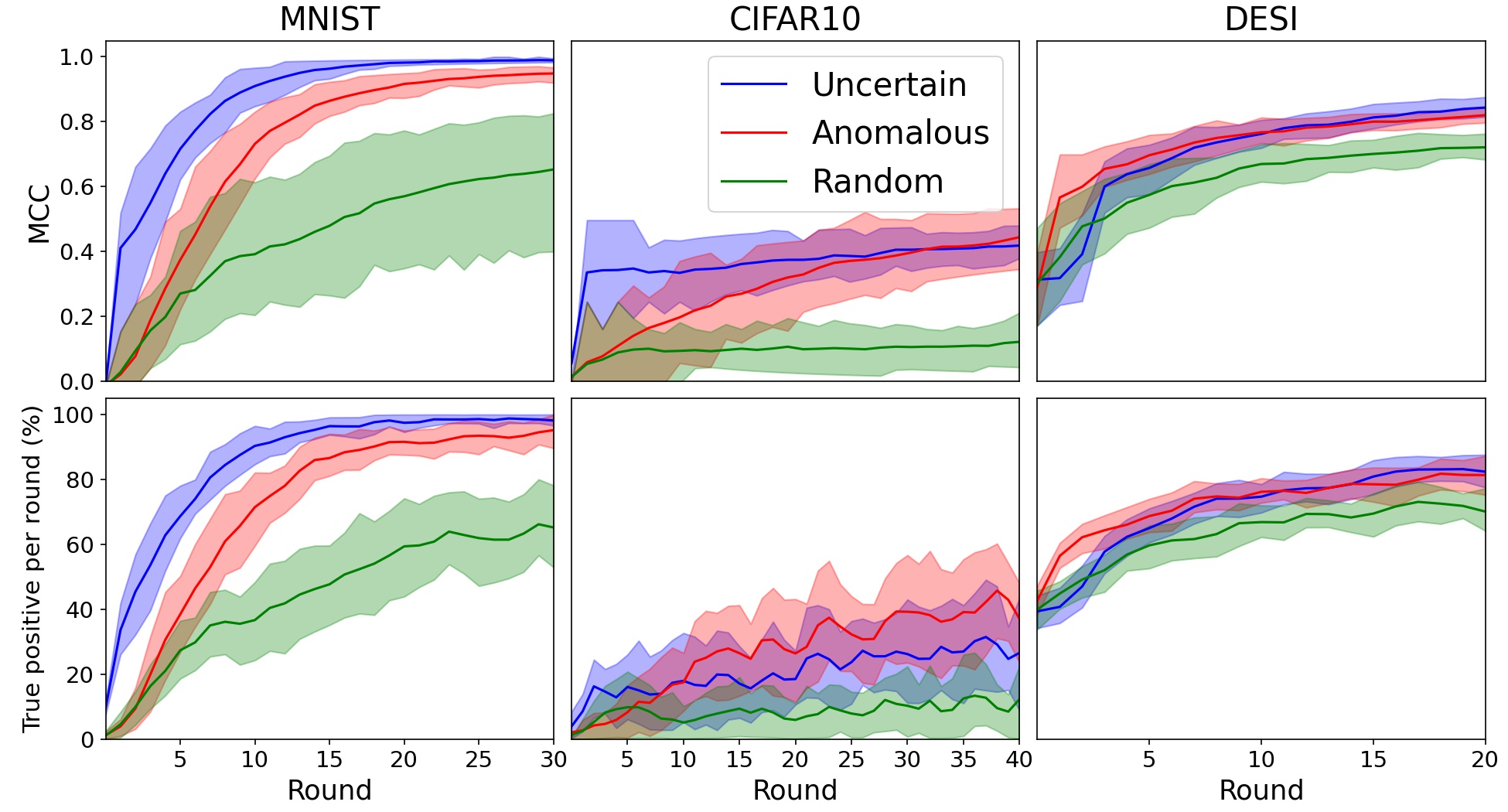}
\caption{Effect of the three different Oracle question selection strategies on the performance of \name on the three datasets MNIST, CIFAR10 and DESI.   
The three question strategies considered are (1) sending the most uncertain data to the Oracle, (2) sending the most anomalous data to the Oracle and (3) sending random data to the Oracle. We see that none of the strategies is optimal in all cases and for all rounds. Of the three, random performed consistently worse that both the Uncertainty and Anomalous strategies, showing that active learning provides a significant performance boost, especially when anomalies are rare relative to the normal classes each round. In the case of DESI, anomalies were a much larger percentage ($\sim 13\%$ of each round of data) than in the other cases, narrowing the advantage provided by the active learning. 
}
\centering
\label{fig:all_three_datasets_Oracle}
\end{figure*}

In the second test of \name we compare it against a model where the full CNN network can learn at every round (not just the last layer as in the previous section), but the Oracle is only given random examples to labels: there is no intelligence in selecting questions. The results are shown in Fig. (\ref{fig:all_three_datasets_Oracle}), where we compare two versions of \namefs; where we select questions for the Oracle based on the two active learning strategies (most uncertain and most anomalous) and compare it to a random selection of questions.

We see that the active learning strategies outperform random selection in all cases, but that the margin of improvement varies significantly depending on the fraction of anomalies in the data at each round. In the case of MNIST and CIFAR-10, where the anomalies are rare (1-2\%) random selection performs poorly, while in the case of DESI, where we chose anomalies to be more common ($\sim11\%$), active learning gives a smaller advantage over random selection. 

More subtle is the comparison of active learning methods of selecting Oracle questions based on most anomalous (highest probability of being in the anomaly class) vs. most uncertain (highest uncertainty index; Eq. \ref{eq:uncertainty_index}). We see that depending on the dataset selecting the most anomalous or the most uncertain examples may yield better improvements, and that the best approach can change over the rounds of active learning.




\begin{table}
\centering
\caption{Average MCC scores for the last round including Active strategy for Dynamic vs Static results (transfer learning) on all detests, showing that allowing the latent space to evolve and adapt dynamically yields significant advantages; see also Fig. \ref{fig:all_three_datasets_dynamic_static_new_2}. Averages are computed over 50 random trials of the entire pipeline.}
\label{table:AllActive}
\begin{tabular}{cccc}
\toprule
{} &    MNIST & CIFAR-10 &  DESI\\
\midrule
Ahunt (Dynamic) &  {\bf 0.99} & {\bf 0.44} & {\bf 0.84} \\
Static  &  0.43 & 0.04 & 0.40 \\
\bottomrule
\end{tabular}
\end{table}

\subsection{\name vs Anomaly Detection Algorithms}

\begin{figure*}
\includegraphics[width=17cm]{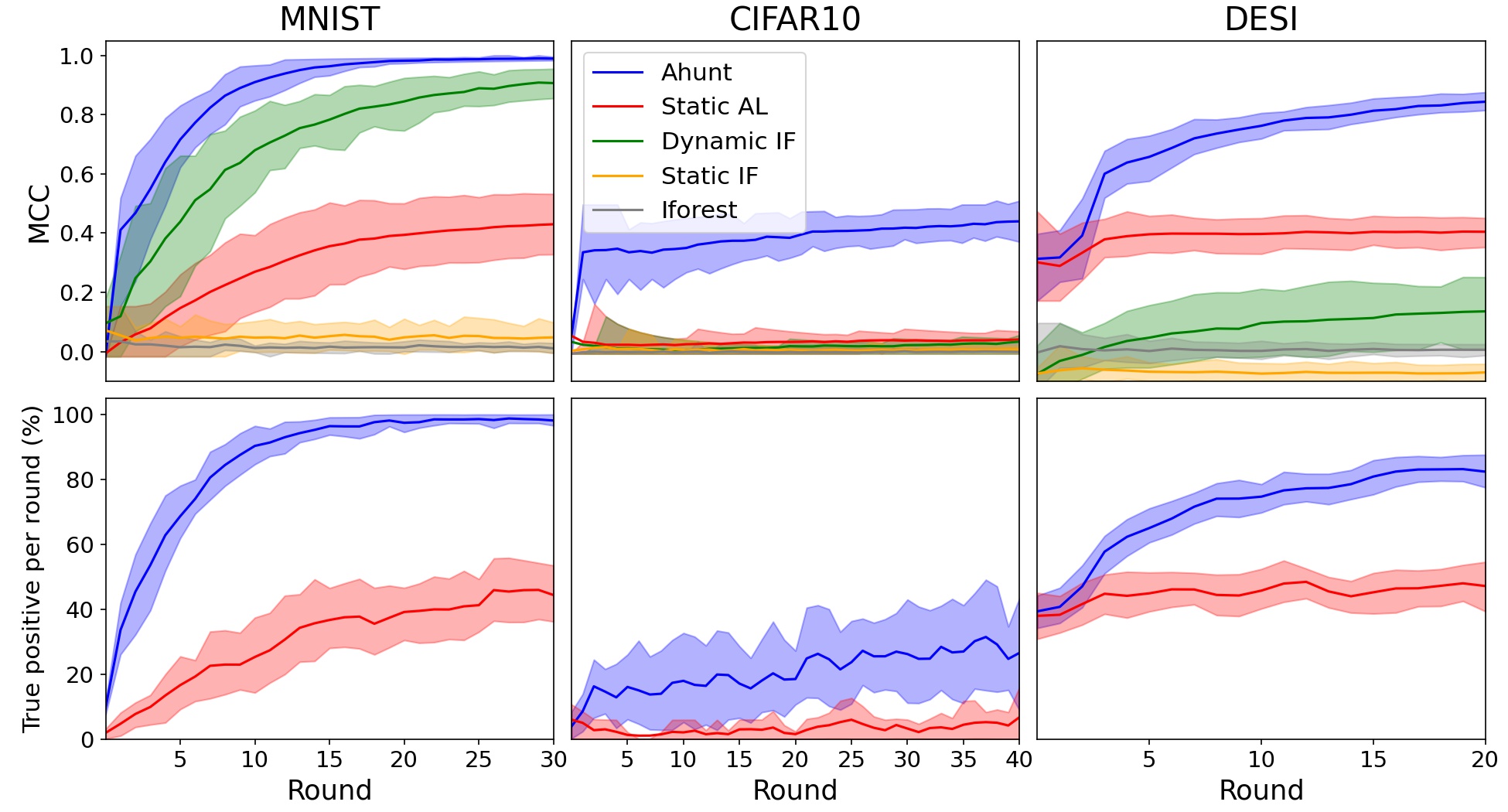}
\centering
\label{fig:all_three_datasets_ALL_Algs}
\caption{Comparison of \name against all algorithms. We see that \name outperforms all algorithms.  Static AL is active learning with only transfer learning. Dynamic IF is Iforest applied to the dynamic latent space. Static IF is Iforest applied to the latent space frozen after round 0 while Iforest is the algorithm applied to the raw input data, which is therefore static by construction. Dynamic latent space (both for \name and Dynamic Iforest (IF) and Active Learning) adds significant value. The 2nd row shows the true positive fraction each round for the algorithms that employ Active Learning. We see that the dynamic latent space (blue curves) significantly improves performance over static latent space (red curves).}

\end{figure*}

A further natural next test of \name is how it performs in comparison with traditional anomaly detection methods; see discussion in section \ref{sec:comparison}. In order to perform this comparison we pick a very popular anomaly detection algorithm: Isolation Forest (Iforest) \citep{liu2008isolation}. We can compare \name to Iforest applied either to the raw input data points or to the outputs of the latent space (the second last layer of the CNN), utilizing our deep learning model as a feature extraction procedure to which we then apply anomaly detection. The second scenario helps us to understand if one can only use the feature extraction part of the deep learning model instead of employing the deep model as an anomaly detection model. We compare MCC for all datasets in Table \ref{table:4}. We see that \name outperforms Iforest in all cases, but that Iforest applied to the latent space is significantly better than applying it directly to the raw input data. This is expected and echos our earlier finding that the dynamic latent space learns good features for anomaly detection.  The full set of comparisons are shown in Fig. (\ref{fig:all_three_datasets_ALL_Algs}) showing all combinations of Iforest (IF).   

We now consider the case where there are multiple unknown anomaly classes in the data and the user's interest in the different anomaly classes 

\begin{table}
	\centering
	\caption{Comparison of MCC scores for \name with isolation forest (Iforest) run on raw input data and the features of the latent space (the second last layer) of the deep convolutional network of the latent space. \name outperforms Iforest in all cases.  The values are reported for the last \OracleEpoch and averaged over 50 randomised trials. }
	\label{table:4}
    \begin{tabular}{cccc}
    \toprule
    {} & Iforest-raw & Iforest-latent &     Ahunt \\
    \midrule
    MNIST   &    0.01 &       0.91 &  {\bf 0.99} \\
    CIFAR10 &    0.00 &       0.03 &  {\bf  0.44} \\
    DESI    &    0.01 &       0.14 &  {\bf  0.84} \\
    \bottomrule
    \end{tabular}
\end{table}


\section{Growing Class Taxonomy \& Changing Interests}\label{sec:dynamic_taxonomy}

If we consider applying \name in real-world scenarios, we must allow for the user's interest in different anomaly classes to change over time as new anomaly classes are discovered. As discussed previously \name requires the user to provide an ``interest" weight vector over the existing classes and the reserve class. This guides the active learning to focus on improving performance on the classes of most interest to the user. Then  when the Oracle, who is typically also the user, confirms a new anomaly class, the user must update their "attention" weight vector to decide how interesting each of the anomaly classes now is. 

To illustrate the flexibility of \name to deal with this scenario we run the following MNIST simulation where there is only one anomaly class present (represented by the ``5" class) initially, in addition to the normal classes. However, after round 10 a second anomaly class (the "9" class)  also begins to appear in the data alongside the first anomaly class. In this scenario, the active learning algorithm needs to modify its attention over the two anomaly and reserve classes dynamically during the observation rounds. 

For the first 10 rounds the data splits between the classes are as the prior MNIST simulation  (see Table \ref{table_data}), i.e. there are six `5's per round. Then, from rounds 10 to 30 there are also ten `9's, representing a second anomaly class. 
In this simulation we have $n_q = 10$, ten questions are put to the Oracle at each round. 

Initially the user sets 100\% of interest on the reserve class (since there are not yet any anomalies that have been observed). After the first anomaly class (`5') is found, the user sets the weight vector to (0.83,0.17) for the reserve class and the `5'-anomaly class respectively. Hence the user is still most interested in finding new classes of anomalies.  As soon as the second anomaly class (`9') is discovered, the user modifies their interest weight vector to (0.125, 0.75, 0.125), for the first anomaly (5), second anomaly class (9's) and the reserve class respectively. This reflects an example where the user is most interested in efficiently detecting the 2nd class of anomalies. 

\begin{figure}
\includegraphics[width=7cm]{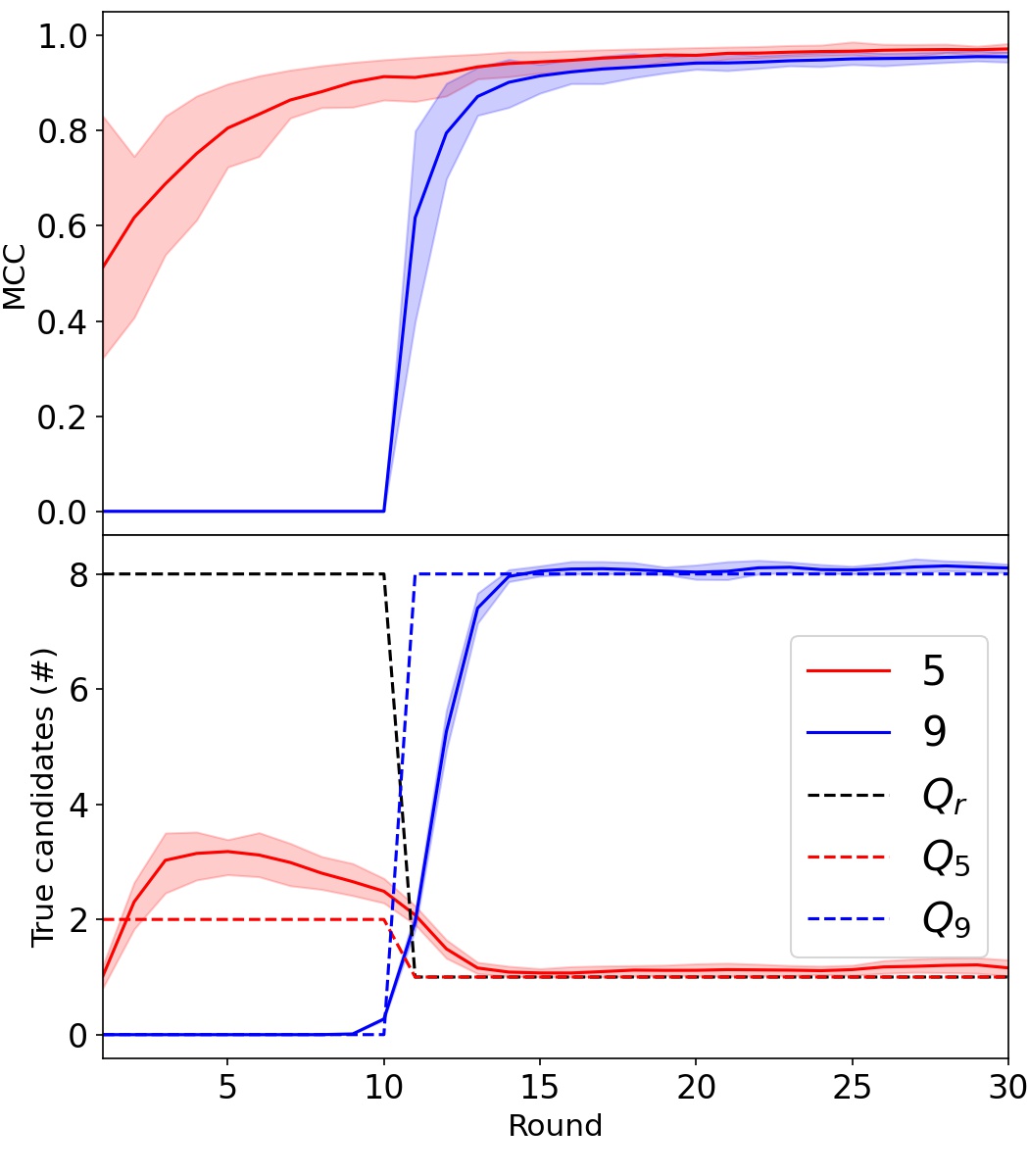}
\caption{
This figure shows how the algorithm can adapt to changing priorities as well as a changing class taxonomy. The Oracle initially has maximum interest on the reserve class
until the first anomaly is found. Then interest is split (0.83, 0.17) between the reserve class and the newly discovered (1st) anomaly class. As soon as the second anomaly class is discovered the user splits their interest (0.125, 0.75, 0.125) over the reserve, 1st anomaly class, 2nd anomaly class and reserve class respectively. The top panel shows total MCC score while the bottom panel shows the number of allowed questions for each class, where the total number of questions is 10. We see that the MCC for the 2nd anomaly class grows very rapidly consistent with the high user interest in that class.}
\label{fig:multi_anomalies}
\centering
\end{figure}

The performance of \name in this scenario is shown in Fig. (\ref{fig:multi_anomalies}).   The upper panel shows the MCC score over the entire dataset for each class over the entire data set. We see that while the MCC for the 1st anomaly class (red) improves quite slowly, since most of the attention is on the reserve class, the MCC for the 2nd class (blue) increases very rapidly since most of the attention is focussed on that class.  The lower panel shows the number of allowed questions for each class at each round (dashed lines) and the percentage of the true questions that correctly identified each class in each round. The results are averaged over $50$ trials with 68\% confidence bands shown.

\section{Conclusions and Future Work}

In this paper we have extended traditional anomaly detection algorithms to allow dynamical evolution of the feature space to facilitate efficient detection of interesting anomalies. This is achieved by using active learning (where a human Oracle is given select examples to label) to intelligently select examples which are then augmented and used to retrain a deep neural network after every active learning round. This leads to the training of a deep network that is progressively optimised to detect the anomaly classes of interest. 

We have shown through extensive ablation tests and comparisons on MNIST, CIFAR-10 and the DESI galaxy image dataset (DESI) that the resulting adaptive feature/latent space provides significant performance enhancements over standard anomaly detection algorithms and active learning applied to a static feature space in general: the algorithm learns better representations of the anomalies, making efficient detection of anomalies easier.

An attractive feature of the new algorithm is that it naturally allows for a changing class structure and taxonomy. Over time the number of known classes will expand as the Oracle  assigns objects to new classes. The user/Oracle is required to specify how interested they are in each of the known classes (including known anomaly classes) and ``unknown unknowns" - the yet undiscovered anomaly classes in the data. This choice of relative interest over classes allows dynamic, personalised anomaly detection that fits each user's interest profile. As a side note, this is a slightly different approach than the one taken in \astronomaly where the user is asked to score (from 1 to 5) every example presented to them in terms of how interesting they are. This allows the problem to be formulated as a regression problem but does not, by itself, allow a taxonomy for the results.

The version of \name that we have presented here is built around an initial phase of supervised deep learning in which the algorithm learns features suitable for efficiently classifying normal classes. This is likely to be very standard, especially in astronomy. What happens if the user only has a large set of ``normal" data to start with, without a set of more nuanced class labels? In this case one could train an autoencoder to learn good encoder features at round 0 for the normal class. Once anomalies have been discovered then the procedure presented here could be used with three classes: `normal, `anomaly' and `reserve'. In this paper we have focused on image anomalies. However the extension to time-series and other types of data is straightforward. 

We end by pointing out that there are intrinsic limitations to any active learning anomaly detection and therefore also to \namefs. If the initial trained features set are such that the Oracle is never shown an anomaly, then the active learning is never able to kick in and begin guiding the feature evolution. 
Future work will look at improved methods for augmenting anomaly data to guide the algorithm. 

\section*{Acknowledgements}

BB thanks Michelle Lochner for many useful discussions over the years. We thank Everlyn Asiko for comments on the draft. Many of the numerical computations were carried out on SARAO facilities, the Baobab and Yggdrasil cluster at University of Geneva and the CHPC.

\section*{Data Availability}

The data used in this paper is available upon request. The code repository will be made public upon acceptance of the paper. 

\appendix
\section{Choice of Loss Function}
\label{sec:loss_function}

In this appendix we explore the impact of choice of loss function on the performance of \namefs. Different loss functions can lead to different amounts of tightness of clustering of the known classes, which in turn can make anomaly detection easier or more difficult.

\begin{figure*} 
\includegraphics[width=17cm]{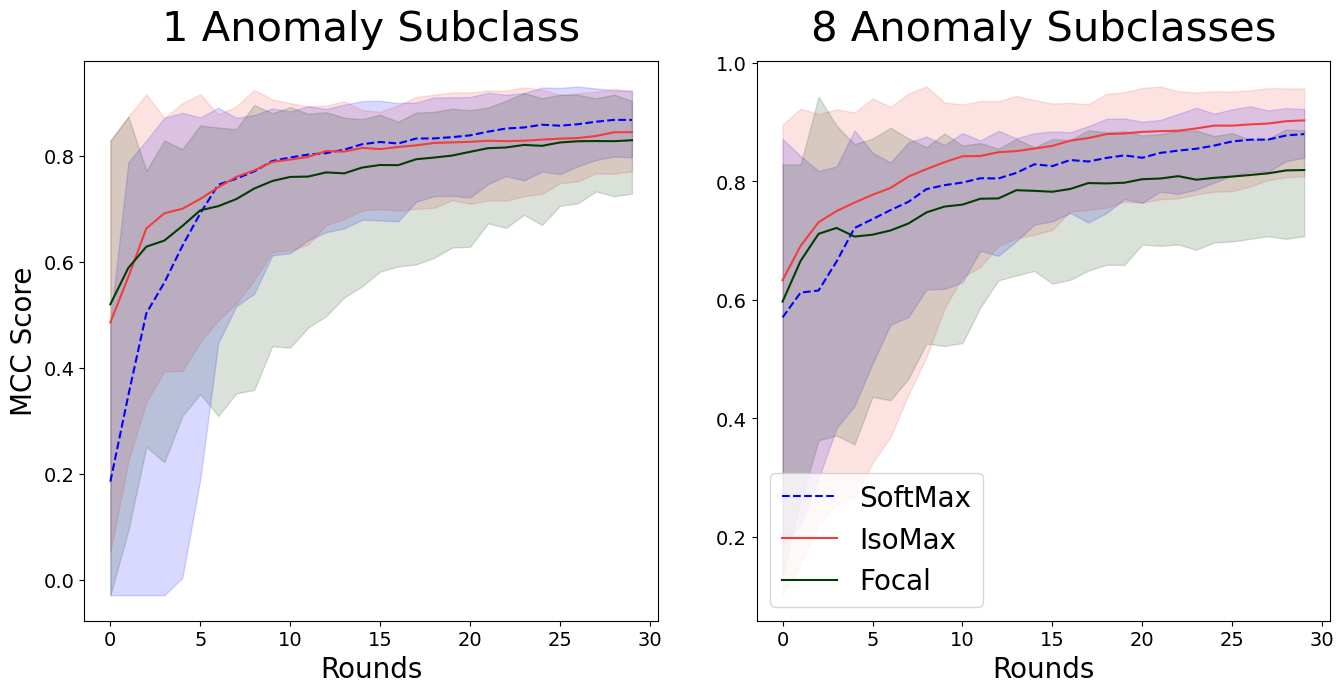}
\caption{Comparison of Loss Functions for CIFAR-10: SoftMax vs Focal vs IsoMax. We see that in the highly heterogeneous case with 8 anomaly subclasses combined into a single anomaly class that IsoMax provides a small boost to MCC score compared to the other loss functions. }
\label{fig:loss_functions}
\centering
\end{figure*}

We compare three loss functions: (1) the standard SoftMax \citep{SoftMax}, (2) the Focal loss function \citep{focalLoss} and (3) the IsoMax loss function. 
The Focal loss function was designed specifically to help deal with situations with large class imbalance, as typically occurs for anomaly detection. This is achieved by assigning a higher weight to hard-to-classify examples. 

The IsoMax loss function was designed specifically for anomaly detection \citep{macedo2019distinction}. Intuitively it works by increasing the intra-class compactness and inter-class separability of features in the latent space. It circumvents the SoftMax loss anistropy by forcing its logits to exclusively depend on the distances between the class prototypes and the learned high-level features.  

We  explore the impact of the loss function on the CIFAR-10 data in two scenarios: (a) the case in which only one of the CIFAR classes is chosen as an anomaly (the ``dog" class) and (b) the case in which 8 classes are lumped together to form a highly heterogeneous anomaly group. In the latter case there is of course much more variability in the anomaly types than when there is only a single anomaly class. In both cases we again have two normal classes.

Fig. (\ref{fig:loss_functions}) shows our results. There is no statistically significant difference between the performance of the loss functions when there is a single class of anomalies. 
However, in the heterogeneous case (where there are 8 anomaly subclasses combined into one), the IsoMax loss  outperforms the SoftMax loss both on  average and in the best case.  This is expected since in this case, there will be several prototypes produced in the feature space and the model's ability to keep the learned features compact will influence the ease of detecting anomalies. In other words, there will be less overlap between the different subclasses in the feature space due to the compactness introduced by the IsoMax loss.

In both cases the Focal loss performs the worst on average, though the difference is not statistically significant relative to the fluctuations from run-to-run. In our setup, we balance the anomaly class by up-sampling it using data augmentation techniques. This helps get around the class imbalance problem and hence the full effect of the focal loss isn't shown. An interesting experiment will be to test the loss functions in a case where the training setup is done without any up-sampling of the minority class. In that case, we expect the focal loss to exhibit better performance relative to the others.



\bibliographystyle{rasti}
\bibliography{ref} 

\begin{thebibliography}{}
\makeatletter
\relax
\def\mn@urlcharsother{\let\do\@makeother \do\$\do\&\do\#\do\^\do\_\do\%\do\~}
\def\mn@doi{\begingroup\mn@urlcharsother \@ifnextchar [ {\mn@doi@}
  {\mn@doi@[]}}
\def\mn@doi@[#1]#2{\def\@tempa{#1}\ifx\@tempa\@empty \href
  {http://dx.doi.org/#2} {doi:#2}\else \href {http://dx.doi.org/#2} {#1}\fi
  \endgroup}
\def\mn@eprint#1#2{\mn@eprint@#1:#2::\@nil}
\def\mn@eprint@arXiv#1{\href {http://arxiv.org/abs/#1} {{\tt arXiv:#1}}}
\def\mn@eprint@dblp#1{\href {http://dblp.uni-trier.de/rec/bibtex/#1.xml}
  {dblp:#1}}
\def\mn@eprint@#1:#2:#3:#4\@nil{\def\@tempa {#1}\def\@tempb {#2}\def\@tempc
  {#3}\ifx \@tempc \@empty \let \@tempc \@tempb \let \@tempb \@tempa \fi \ifx
  \@tempb \@empty \def\@tempb {arXiv}\fi \@ifundefined
  {mn@eprint@\@tempb}{\@tempb:\@tempc}{\expandafter \expandafter \csname
  mn@eprint@\@tempb\endcsname \expandafter{\@tempc}}}

\bibitem[\protect\citeauthoryear{Aggarwal et~al.}{Aggarwal
  et~al.}{2018}]{aggarwal2018neural}
Aggarwal C.~C.,  et~al., 2018, Springer, 10, 978

\bibitem[\protect\citeauthoryear{Basu, Banerjee  \& Mooney}{Basu
  et~al.}{2004}]{activelearning-01}
Basu S.,  Banerjee A.,   Mooney R.,  2004, in Berry M.,  Dayal U.,  Kamath C.,
   Skillicorn D.,  eds, Proceedings of the Fourth SIAM International Conference
  on Data Mining. pp 333--344

\bibitem[\protect\citeauthoryear{Deng}{Deng}{2012}]{deng2012mnist}
Deng L.,  2012, IEEE Signal Processing Magazine, 29, 141

\bibitem[\protect\citeauthoryear{Dey et~al.,}{Dey
  et~al.}{2019}]{dey2019overview}
Dey A.,  et~al., 2019, The Astronomical Journal, 157, 168

\bibitem[\protect\citeauthoryear{Fortson, Masters, Nichol, Edmondson, Lintott,
  Raddick  \& Wallin}{Fortson et~al.}{2012}]{fortson2012galaxy}
Fortson L.,  Masters K.,  Nichol R.,  Edmondson E.,  Lintott C.,  Raddick J.,
  Wallin J.,  2012, Advances in machine learning and data mining for astronomy,
  2012, 213

\bibitem[\protect\citeauthoryear{Hložek et~al.,}{Hložek
  et~al.}{2020}]{plasticc}
Hložek R.,  et~al., 2020, Results of the Photometric LSST Astronomical
  Time-series Classification Challenge (PLAsTiCC),
  \mn@doi{10.48550/ARXIV.2012.12392}, \url {https://arxiv.org/abs/2012.12392}

\bibitem[\protect\citeauthoryear{Krizhevsky, Hinton  et~al.}{Krizhevsky
  et~al.}{2009}]{krizhevsky2009learning}
Krizhevsky A.,  Hinton G.,   et~al., 2009, Citeseer

\bibitem[\protect\citeauthoryear{Lin, Goyal, Girshick, He  \& Doll{\'{a}}r}{Lin
  et~al.}{2017}]{focalLoss}
Lin T.,  Goyal P.,  Girshick R.~B.,  He K.,   Doll{\'{a}}r P.,  2017, \mn@doi
  [CoRR] {10.48550/ARXIV.1708.02002}, abs/1708.02002

\bibitem[\protect\citeauthoryear{Liu, Ting  \& Zhou}{Liu
  et~al.}{2008}]{liu2008isolation}
Liu F.~T.,  Ting K.~M.,   Zhou Z.-H.,  2008, in 2008 eighth ieee international
  conference on data mining. pp 413--422

\bibitem[\protect\citeauthoryear{Liu, Wen, Yu  \& Yang}{Liu
  et~al.}{2016}]{SoftMax}
Liu W.,  Wen Y.,  Yu Z.,   Yang M.,  2016, in Proceedings of the 33rd
  International Conference on International Conference on Machine Learning -
  Volume 48. ICML'16.
JMLR.org, p. 507–516

\bibitem[\protect\citeauthoryear{Lochner \& Bassett}{Lochner \&
  Bassett}{2021}]{lochner2021astronomaly}
Lochner M.,  Bassett B.~A.,  2021, Astronomy and Computing, 36, 100481

\bibitem[\protect\citeauthoryear{Lochner \& Bassett}{Lochner \&
  Bassett}{2022}]{astronomaly_hitch}
Lochner M.,  Bassett B.~A.,  2022, A Hitchhiker's Guide to Anomaly Detection
  with Astronomaly, \mn@doi{10.48550/ARXIV.2201.10189}, \url
  {https://arxiv.org/abs/2201.10189}

\bibitem[\protect\citeauthoryear{Lochner et~al.,}{Lochner
  et~al.}{2022}]{lsst_optimal}
Lochner M.,  et~al., 2022, The Astrophysical Journal Supplement Series, 259, 58

\bibitem[\protect\citeauthoryear{Mac{\^e}do, Ren, Zanchettin, Oliveira, Tapp
  \& Ludermir}{Mac{\^e}do et~al.}{2019}]{macedo2019distinction}
Mac{\^e}do D.,  Ren T.~I.,  Zanchettin C.,  Oliveira A.~L.,  Tapp A.,
  Ludermir T.,  2019, arXiv preprint arXiv:1908.05569

\bibitem[\protect\citeauthoryear{Matthews}{Matthews}{1975}]{matthews1975comparison}
Matthews B.~W.,  1975, Biochimica et Biophysica Acta (BBA)-Protein Structure,
  405, 442

\bibitem[\protect\citeauthoryear{Nixon, Sedky  \& Hassan}{Nixon
  et~al.}{2021}]{nixon2021salad}
Nixon C.,  Sedky M.,   Hassan M.,  2021, \mn@doi [TechRxiv]
  {10.36227/techrxiv.14896773.v1}

\bibitem[\protect\citeauthoryear{Pelleg \& Moore}{Pelleg \&
  Moore}{2004}]{pelleg2004active}
Pelleg D.,  Moore A.,  2004, Advances in neural information processing systems,
  17, 1073

\bibitem[\protect\citeauthoryear{Pimentel, Monteiro, Viana, Veloso  \&
  Ziviani}{Pimentel et~al.}{2018}]{pimentel2018generalized}
Pimentel T.,  Monteiro M.,  Viana J.,  Veloso A.,   Ziviani N.,  2018, stat,
  1050, 23

\bibitem[\protect\citeauthoryear{Roberts, Bassett  \& Lochner}{Roberts
  et~al.}{2020}]{badac}
Roberts E.,  Bassett B.,   Lochner M.,  2020, International Journal of Hybrid
  Intelligent Systems, vol. 16, pp 207--222

\bibitem[\protect\citeauthoryear{Sadr, Bassett  \& Kunz}{Sadr
  et~al.}{2021}]{sadr2021flexible}
Sadr A.~V.,  Bassett B.~A.,   Kunz M.,  2021, Neural Computing and
  Applications, pp 1--11

\bibitem[\protect\citeauthoryear{Walmsley et~al.,}{Walmsley
  et~al.}{2022}]{Walmsley_2022}
Walmsley M.,  et~al., 2022, \mn@doi [Monthly Notices of the Royal Astronomical
  Society] {10.1093/mnras/stac525}, 513, 1581–1599

\bibitem[\protect\citeauthoryear{Webb et~al.,}{Webb
  et~al.}{2020}]{astronomaly_apply}
Webb S.,  et~al., 2020, Monthly Notices of the Royal Astronomical Society, 498,
  3077

\bibitem[\protect\citeauthoryear{Wolpert \& Macready}{Wolpert \&
  Macready}{1997}]{NFL}
Wolpert D.,  Macready W.,  1997, \mn@doi [IEEE Transactions on Evolutionary
  Computation] {10.1109/4235.585893}, 1, 67

\makeatother
\end{thebibliography}





\bsp	
\label{lastpage}
\end{document}